\documentclass{article}
\usepackage{multirow}
\usepackage{booktabs}
\usepackage{spconf,amsmath,graphicx}
\usepackage{url}
\usepackage{todonotes}

\title{IMPROVING SINGING VOICE SEPARATION WITH THE WAVE-U-NET \\ USING MINIMUM HYPERSPHERICAL ENERGY}

\name{Joaquin Perez-Lapillo, Oleksandr Galkin, Tillman Weyde}
\address{Department of Computer Science \\
City, University of London \\
\texttt{\{joaquin.perez-lapillo,oleksandr.galkin,t.e.weyde\}@city.ac.uk}}

\begin{document}

\maketitle

\begin{abstract}
In recent years, deep learning has surpassed  traditional approaches to the problem of singing voice separation.
The Wave-U-Net is a  recent deep network architecture that operates directly on the time domain. 
The standard Wave-U-Net is trained with data augmentation and early stopping to prevent overfitting. 
Minimum hyperspherical energy (MHE)  regularization has recently proven to increase generalization in image classification problems by encouraging a diversified filter configuration.
In this work, we apply MHE regularization to the 1D filters of the Wave-U-Net.
We evaluated this approach for separating the  vocal part from mixed music audio recordings on the MUSDB18 dataset. 
We found that adding MHE regularization to the loss function  consistently improves singing voice separation, as measured in the Signal to Distortion Ratio on test recordings, leading to the current best time-domain system for singing voice extraction.  
\end{abstract}

\begin{keywords}
 Minimum Hyperspherical Energy, Music Source Separation, Deep Learning, Regularization 
\end{keywords}

\section{Introduction}
\label{sec:intro}

Audio source separation is a core research area in audio signal processing with applications in speech, music and environmental audio. 
The main goal is to extract one or more target sources while suppressing other sources and noise \cite{vincent_audio_2018}. 
One case is the separation of a singing voice, commonly presenting the main melody, from the musical accompaniment. 
Separating vocals from accompaniment has several applications, 
such as editing and remixing music, automatic transcription, generating karaoke tracks and music information retrieval \cite{nugraha_multichannel_2016}.

We address this problem by applying Minimum Hyperspherical Energy (MHE) regularization to the Wave-U-Net model. In particular our contributions are:
\begin{itemize}
    \item A novel application of MHE to 1D convolutional neural networks with open source implementation. 
    \item An extensive evaluation of MHE configurations and hyperparameters for singing voice separation.
    \item A new state-of-the-art for time-domain singing voice extraction on MUSDB18.
\end{itemize}

\section{Related Work}
\label{sec:related work}

Approaches to source separation can be broadly divided into two groups: traditional and deep neural networks \cite{rafii_overview_2018}. 
The first group contains techniques such as non-negative matrix factorization, Bayesian methods, and the analysis of repeating structures \cite{smaragdis_static_2014,ozerov_adaptation_2007, rafii_repeating_2013}.
However, several deep neural network architectures have surpassed those models and achieve state-of-the-art performance today.
\cite{jansson_singing_2017} introduced the U-Net architecture for singing voice extraction, using spectrograms. 
The Wave-U-Net, proposed by \cite{stoller_wave-u-net:_2018}, processes the audio data in the time domain, and is trained as a deep \textit{end-to-end} model with state-of-the-art performance.

Many strategies for regularization have been developed, with early stopping, data augmentation, weight decay and dropout
being the most commonly used \cite{Goodfellow-et-al-2016}. 
Neural network based singing voice separation systems have used some of these methods in the past.
\cite{jansson_singing_2017} uses dropout. 
In \cite{stoller_wave-u-net:_2018}, data augmentation is applied by varying the balance of the sources.
Both methods also use early stopping.

In deep learning in general, overfitting has also been addressed by compressing the network \cite{han_deep_2015}, modifying the network architecture \cite{howard_mobilenets:_2017}, and alleviating redundancy through diversification  \cite{xie_diversity-promoting_2017}.
This last approach enforces diversity between neurons via regularization.
Minimum Hyperspherical Energy, proposed by \cite{liu_learning_2018}, also falls into this category. 
Inspired by a problem in physics \cite{jj_thomson_1904}, the MHE term aims to encourage variety between filters.
MHE has increased performance in image classification, but the method has not yet been applied to audio.

\section{Method}
\label{sec:methodology}

\subsection{Architecture}
Our work is based on the best Wave-U-Net configuration found by \cite{stoller_wave-u-net:_2018} for singing voice separation, which is a deep U-Net composed of 24 hidden convolutional layers, 12 in each path, considering a filter size of 15 in the downsampling path and 5 for the upsampling. 
The first layer computes 24 feature maps, and each subsequent layer doubles that number. 
Each layer in the upsampling path is also connected to their corresponding layer in the downsampling path with skip-connections.
The output layer generates an estimated waveform in the range [-1,1] for the vocals, while the accompaniment is obtained as the difference between the original mixture and the estimated singing voice, as in \cite{stoller_wave-u-net:_2018}. 
The original loss function is mean squared error (MSE) per sample.

\subsection{Adding MHE to Wave-U-Net training}

The MHE regularised loss function for the Wave-U-Net can be defined as: \begin{equation}
    Loss = 
    MSE
    + \lambda_h * \sum_{j=1}^{L} \frac{1}{N_j(N_j-1)} { E_s }_j,
    \label{eqn:loss}
\end{equation} 
where $L$ is the number of hidden layers and $N_j$ is the number of filters in the $j$-th layer. 
$E_{sj}$ represents the hyperspherical energy of the $j$-th layer, given the parameter $s$, and it is calculated as:
\begin{equation}
    { E_s }_j
    = \sum_{i=1}^{N_j} \sum_{k=1, k \neq i}^{N_j}
    f_s(||\hat{w}_i - \hat{w}_k ||), 
    \label{eqn:mhe_energy}
\end{equation}
where $||*||$  is the Euclidean norm and $\hat{w}_i = \frac{w_i}{||w_i||}$ is the weight vector of neurons $i$ projected onto a unit hypersphere. 
The dimensionality of the hypersphere is given by the number of incoming weights per neuron.
For $f_s$, we use $f_s(z) = z^{-s}$  for $s>0$, and $f_s(z) = \log z^{-1}$ otherwise \cite{liu_learning_2018}.

With regard to $\lambda_h$, i.e. the weighting constant of the regulariser, \cite{liu_learning_2018} recommends to use a constant depending on the number of hidden layers, with the aim of reducing the weighting of MHE for very deep architectures.
We use here $\lambda_h = \frac{1}{L}$.

There are two possible configurations of MHE, full or half space. 
The half-space variation creates a \textit{virtual} neuron for every existing neuron with inverted signs of the weights. 
The second term of equation \ref{eqn:loss} is then applied to $2N_j$ neurons in each hidden layer $j$.
There are two alternative distance functions, Euclidean and angular. 
When dealing with angular distances, $||\hat{w_i} - \hat{w_j}||$ is replaced with $arccos(\hat{w_i}^T \hat{w_j})$.
The parameter \textit{s} controls the scaling of MHE. 
Using the convention of \cite{liu_learning_2018} we have a total of twelve possible configurations as shown in Table~\ref{tab:mhe}.
\begin{table}[tbh]
    \centering
    \caption{MHE configurations. Values of $s$ with prefix $a$ use angular distance. Each model can be used with each $s$ \cite{liu_learning_2018}.}
    \begin{tabular}{|c|c|}
    \hline
    Parameter & Values \\ 
    \hline
    model & [MHE, half\_MHE] \\ 
    \hline
    $s$ value & [0, 1, 2, a0 ,a1, a2] \\ \hline
\end{tabular}
\label{tab:mhe}
\end{table}

\subsection{Dataset}

We use the MUSDB18 dataset \cite{MUSDB18}\footnote{\url{https://sigsep.github.io/datasets/musdb.html}}. 
It is composed of 150 full-length musical tracks. 
The data is divided into 100 songs for development and 50 for testing, corresponding to 6.5 and 3.5 hours, respectively. 
The development set is further divided into training and validation sets by randomly selecting 25 tracks for the latter.

For the task of singing voice separation, the drums, bass, and other sources are mixed to represent the accompaniment. 

For the final experiment, we added the CCMixter dataset \cite{liutkus_scalable_2015} featuring 50 more songs for training, as in \cite{stoller_wave-u-net:_2018}.
In this setup, the last model is trained with a total of 150 songs. 
The test set remains unchanged.

\subsection{Experimental setup}

We explored the influence of the MHE configuration and hyperparameters with an initial grid search. 
For the grid search, all twelve MHE configurations described in Table \ref{tab:mhe} and a baseline model were implemented using the same network parametrization in order to compare their performance.
Our implementation is available as open source\footnote{\url{https://github.com/jperezlapillo/Hyper-Wave-U-Net}}.

As in \cite{stoller_wave-u-net:_2018}, the models were trained using the Adam optimizer \cite{kingma_adam:_2015} with decay rates of $\beta_1 = 0.9$ and $\beta_2 = 0.999$, a batch size of 16 and learning rate of $0.0001$.
Data augmentation is applied by attenuating the vocals by a random factor in the range [0.7,1.0]. 
The best model selected in the first round of training is further fine-tuned, doubling the batch size and reducing the learning rate to $0.00001$. 
The model with the lowest validation loss is finally tested against the unseen data.

In order to reduce the computational cost for the exploratory experiments, an epoch was defined as 1,000 iterations, instead of the original 2,000, and the early stopping criterion was reduced from 20 to 10 epochs without improvement of validation loss. 
Additionally, the tracks were mixed down to mono. 

A further experiment was performed exploring the regularization constant  $\lambda_h=1/L$ and an increased early stopping criterion.
We finally evaluated the singing voice separation performance of the best MHE configuration with the original settings using the extended dataset and the parametrisation as in \cite{stoller_wave-u-net:_2018}.

\subsection{Evaluation}

The models are evaluated using the signal-to-distortion ratio (SDR), as proposed by \cite{vincent_performance_2006}.
The audio tracks are partitioned into several non-overlapping segments of length one second to calculate SDR for each individual segment. 
The resulting SDRs are then averaged over the songs, and over the whole dataset. 

For near-silent parts in one source the SDR can reach very low values, which can affect the global mean statistic. 
To deal with this issue, median statistics are also provided as in \cite{stoller_wave-u-net:_2018}, along with standard deviation and median absolute deviation (MAD) for vocals and accompaniment. 

\section{Results}
\label{sec:results}

\subsection{Hyperparameter exploration}

\begin{figure}[t]
    \centering
    \includegraphics[scale=0.55]{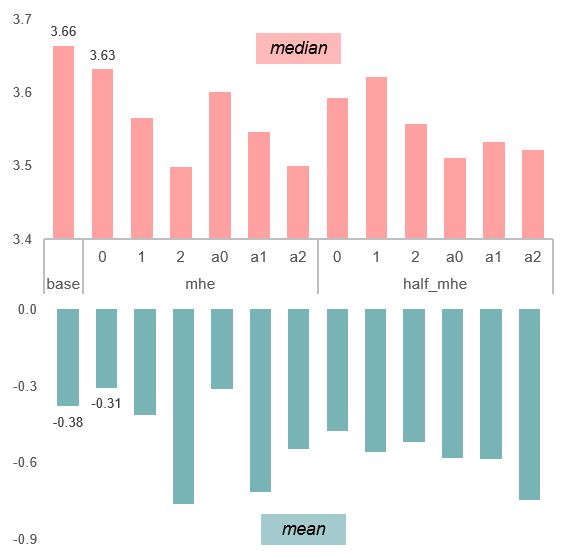}
    \caption{Test set SDR results (in dB) for singing voice.}
    \label{fig:grid_vocals}
\end{figure}

\begin{figure}[t]
    \centering
    \includegraphics[scale=0.56]{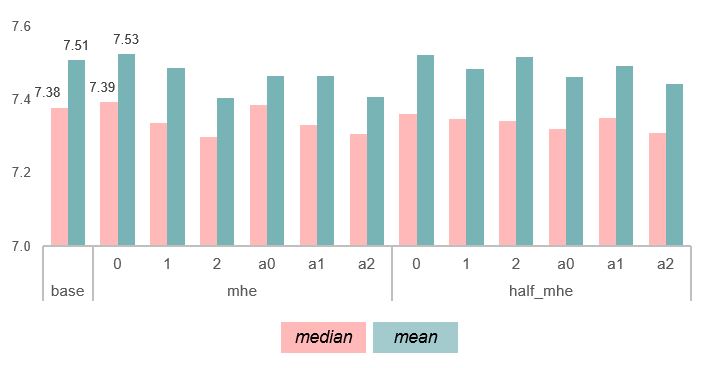}
    \caption{Test set SDR results (in dB) for accompaniment.}
    \label{fig:grid_acc}
\end{figure}

Figures \ref{fig:grid_vocals} and \ref{fig:grid_acc} show statistics for the SDR obtained by each model over the MUSDB18 test set. 
The baseline model obtained the highest median SDR with 3.66 dB for the vocals estimation, followed closely by MHE\_0 (full-space MHE, Euclidean distance, $s=0$,) with 3.63 dB. 
However, MHE\_0 obtained the highest mean SDR with -0.31 dB, compared to -0.38 dB for the baseline.
For the accompaniment, the MHE\_0 model showed the highest median and mean SDR, albeit by a small margin.

\begin{table*}[tb]
  \centering
  \caption{Average test set SDR results (in dB) aggregated by MHE  hyperparameters.}
  \begin{tabular}{|cc|cc|ccc|cc|}
    \hline
          &       & \multicolumn{2}{c|}{Model} & \multicolumn{3}{c|}{$s$ value} & \multicolumn{2}{c|}{Distance} \\
          &       & mhe   & half\_mhe & 0     & 1     & 2     & euclidean & angular \\
    \hline
    \multirow{4}[2]{*}{Vocals} & \multicolumn{1}{l|}{Med} & 3.56  & 3.56  & \textbf{3.58} & 3.57  & 3.52  & \textbf{3.58} & 3.54 \\
          & \multicolumn{1}{l|}{MAD} & 2.85  & 2.90  & 2.89  & 2.87  & 2.86  & 2.87  & 2.87 \\
          & \multicolumn{1}{l|}{Mean} & \textbf{-0.51} & -0.58 & \textbf{-0.42} & -0.57 & -0.64 & \textbf{-0.51} & -0.58 \\
          & \multicolumn{1}{l|}{SD} & 13.81 & 14.02 & 13.78 & 13.99 & 13.97 & 13.89 & 13.94 \\
    \hline
    \multirow{4}[2]{*}{Accomp.} & \multicolumn{1}{l|}{Med} & 7.34  & 7.34  & \textbf{7.37} & 7.34  & 7.31  & \textbf{7.35} & 7.33 \\
          & \multicolumn{1}{l|}{MAD} & 2.11  & 2.11  & 2.11  & 2.11  & 2.10  & 2.10  & 2.11 \\
          & \multicolumn{1}{l|}{Mean} & 7.46  & \textbf{7.49} & \textbf{7.49} & 7.48  & 7.44  & \textbf{7.49} & 7.45 \\
          & \multicolumn{1}{l|}{SD} & 3.88  & 3.81  & 3.87  & 3.82  & 3.84  & 3.83  & 3.87 \\
    \hline
  \end{tabular}
  \label{tab:mhe_config}
\end{table*}

Table \ref{tab:mhe_config} compares different MHE configurations, aggregating the results of Figures \ref{fig:grid_vocals} and \ref{fig:grid_acc} by groups of parameters.
From this, it becomes clear that the ideal \textit{s} value is $0$ and the preferred distance is \textit{Euclidean}. 
When comparing MHE versus half\_MHE models, the former obtains better results on vocals, and the latter on the accompaniment estimation.

\subsection{MHE loss curves}

\begin{figure}[t]
    \centering
    \includegraphics[scale=0.32]{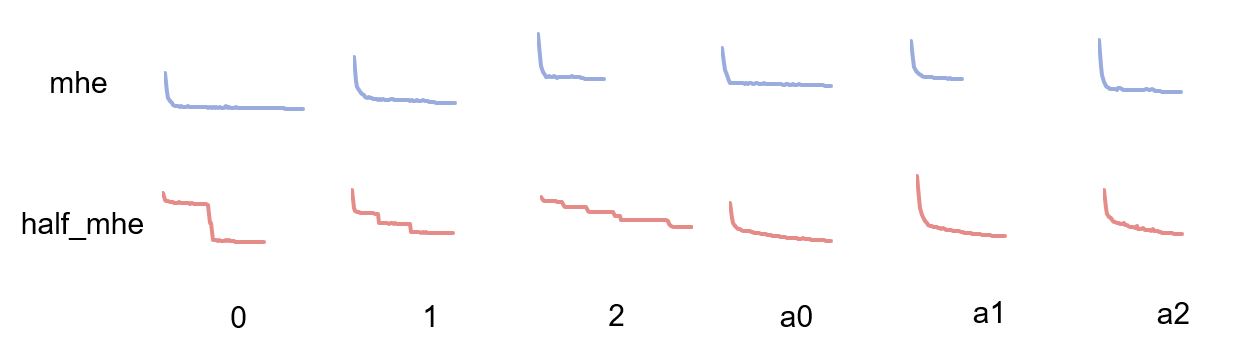}
    \caption{MHE loss dynamics over training process. Longer curves represent longer training processes.}
    \label{fig:dynamics}
\end{figure}

Figure \ref{fig:dynamics} shows the development of the MHE loss during training. 
While full MHE models rapidly reduce the MHE loss in the first epochs of the training process, Euclidean half\_MHE models tend to form steps.
The behaviour seems to increase in frequency when higher \textit{s} values are in use. 

The MHE loss tends to be relatively stable, with changes made every epoch in the range of $[0.0001, 0.00001]$.
This is probably due to the large number of parameters being considered and the value chosen for learning rate ($0.0001$).

The following experiments focus on comparing the baseline model with the best performing MHE model, MHE\_0, in different scenarios.

\subsection{Early stopping and regularization strength}

Table~\ref{tab:sdr_lambda} shows the SDR results for MHE\_0 with 20 epochs early stopping criterion and different values for $\lambda_h$.
This doubling of the early stopping criterion led to a 88\% increase in the number of epochs for the MHE\_0 model.
In this setting the MHE\_0 model obtained higher results in all statistics compared to the baseline model on the leftmost column.

The increased early stopping criterion leads to longer training times for the MHE\_0 model with $\lambda_h = 1/L$. 

So far, the regularization constant $\lambda_h$ was set to $1/L$ following the recommendation in \cite{liu_learning_2018}. 
Table \ref{tab:sdr_lambda} shows the SDR results for $\lambda_h=1/(2L)$ and $\lambda_h=1$, too. 
It is clear that both increasing and decreasing the regularization constant have a negative effect on SDR. 

\begin{table}[htbp]
  \centering
  \caption{Test set SDR (in dB) for 20 epochs early stopping.}
    \begin{tabular}{|cc|c|c|c|c|}
    \hline
           & Value of $\lambda_h$  & Basel. & 1/(2L)  & 1/L  & 1 \\
    \hline
    \multirow{4}[2]{*}{Voc.} & \multicolumn{1}{l|}{Med} & 3.65 & 3.50  & \textbf{3.69} & 3.64 \\
          & \multicolumn{1}{l|}{MAD} & 3.04 & 2.82  & 2.98  & 2.96 \\
          & \multicolumn{1}{l|}{Mean} & -0.56 & -0.39 & \textbf{0.01} & -0.10 \\
          & \multicolumn{1}{l|}{SD} & 14.23 & 13.53 & 13.31 & 13.30 \\
    \hline
    \multirow{4}[2]{*}{Acc.} & \multicolumn{1}{l|}{Med} & 7.37 & 7.32  & \textbf{7.44} & 7.42 \\
          & \multicolumn{1}{l|}{MAD} & 2.10 & 2.11  & 2.14  & 2.14 \\
          & \multicolumn{1}{l|}{Mean} & 7.53 & 7.44  & \textbf{7.56} & 7.54 \\
          & \multicolumn{1}{l|}{SD} & 3.74 & 3.92  & 3.89  & 3.89 \\
    \hline
    \multicolumn{2}{|c|}{Epochs} & 120 & 76  & 167  & 120 \\
    \hline
    \end{tabular}
  \label{tab:sdr_lambda}
\end{table}

\subsection{Extended training data}

Considering the same conditions reported by \cite{stoller_wave-u-net:_2018} for their best vocals separator system, called M4, we re-implemented this configuration and trained it in parallel with an MHE\_0 model. 
The results in Table \ref{tab:sdr_final} show that MHE\_0 outperforms our M4 implementation (left row) and the originally reported M4 results.
This shows that MHE\_0 is currently the best time-domain singing voice separation model.

\begin{table}[htbp]
  \centering
  \caption{Test set SDR (in dB) for the extended training set.}
    \begin{tabular}{|cc|c|c|c|}
    \hline
          &    Model   & M4(reimp.)  & MHE\_0   & M4(orig.) \\
    \hline
    \multirow{4}[2]{*}{Voc.} & \multicolumn{1}{l|}{Med} & 4.44  & \textbf{4.69} & 4.46 \\
          & \multicolumn{1}{l|}{MAD} & 3.15  & 3.24  & 3.21 \\
          & \multicolumn{1}{l|}{Mean} & 0.17 & \textbf{0.75} & 0.65 \\
          & \multicolumn{1}{l|}{SD} & 14.38 & 13.91 & 13.67 \\
    \hline
    \multirow{4}[2]{*}{Acc.} & \multicolumn{1}{l|}{Med} & 10.54  & \textbf{10.88} & 10.69 \\
          & \multicolumn{1}{l|}{MAD} & 3.01  & 3.13  & 3.15 \\
          & \multicolumn{1}{l|}{Mean} & 11.71  & \textbf{12.10} & 11.85 \\
          & \multicolumn{1}{l|}{SD} & 6.48  & 6.77  & 7.03 \\
    \hline
    \multicolumn{2}{|c|}{Epochs} & 90    & 122   & - \\
    \hline
    \end{tabular}
  \label{tab:sdr_final}
\end{table}

\begin{figure}
    \centering
    \includegraphics[width=1.0\columnwidth]{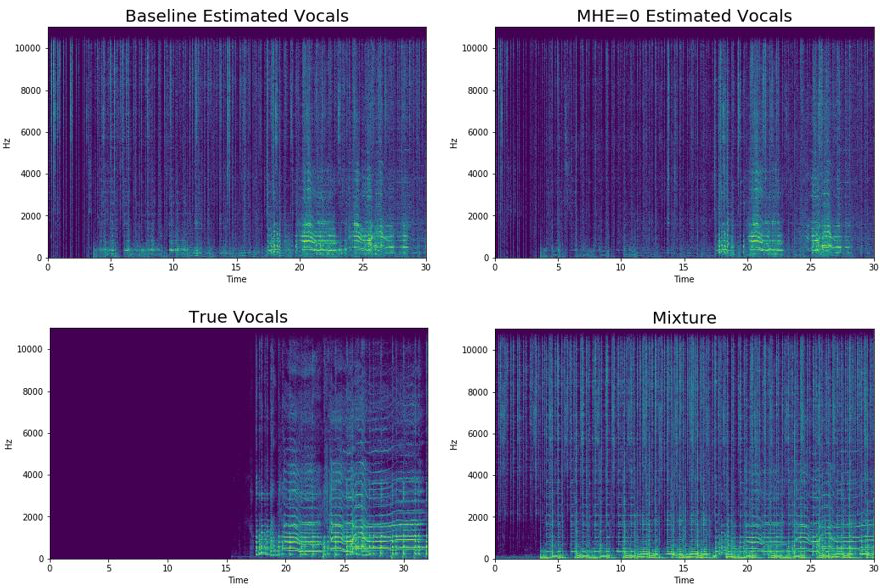}
    \caption{Spectrograms of vocals with silent period.}
    \label{fig:spectro}
\end{figure}

\section{Discussion}
\label{sec:discussion}

The results confirm that the MHE\_0 model, i.e. full-space MHE with Euclidean distance and \textit{s} value$=0$, outperforms the alternative MHE versions, similar to the results in \cite{liu_learning_2018}.
The training for MHE models needs more epochs compared to the baseline and benefits from an increased early stopping criterion. 
Overall, including MHE regularization in the Wave-U-Net loss function improves singing voice separation, when sufficient time is given for the training. 

Particularly in the vocals source, there are some silent periods. 
These can  produce very low SDR results and with very audible artifacts. 
The MHE helps to reduce artifacts in these periods, as can be seen in Figure~\ref{fig:spectro}. 

The results achieved here do not yet match the best singing voice extracting method based on time-frequency representations \cite{stoter_open-unmix_2019}. 
The performance gap of around 1.5 dB is a research challenge that is worthwhile, because of the time-domain systems' potential for low latency processing. 

\section{Conclusion}
\label{sec:conclusion}

This study explores the use of a novel regularization method, minimum hyperspherical energy (MHE), for improving the task of singing voice separation in the Wave-U-Net. 
It is, to the authors' knowledge, the first time that this technique is applied to an audio-related problem. 

Our results suggest that MHE regularization, combined with the appropriate early stopping criterion, is worth including in the loss function of deep learning separator systems such as the Wave-U-Net, as it leads to a new state of the art in our experiments.  

For future work we intend to address other applications, such as speech enhancement and separation, as well as other loss formulations. 
We are also interested in designing low latency systems based on this approach and aim to reduce the computational cost of the Wave-U-Net.

\vfill\pagebreak

\bibliographystyle{IEEEbib}
\bibliography{references}

\end{document}